\documentclass[sigconf]{acmart}

\usepackage{booktabs}
\usepackage{subcaption}
\usepackage{tabularx}
\usepackage{xcolor}
\usepackage{xurl}

\newcommand{\FigProblem}{%
\begin{figure*}[t]
\centering
\includegraphics[width=\textwidth]{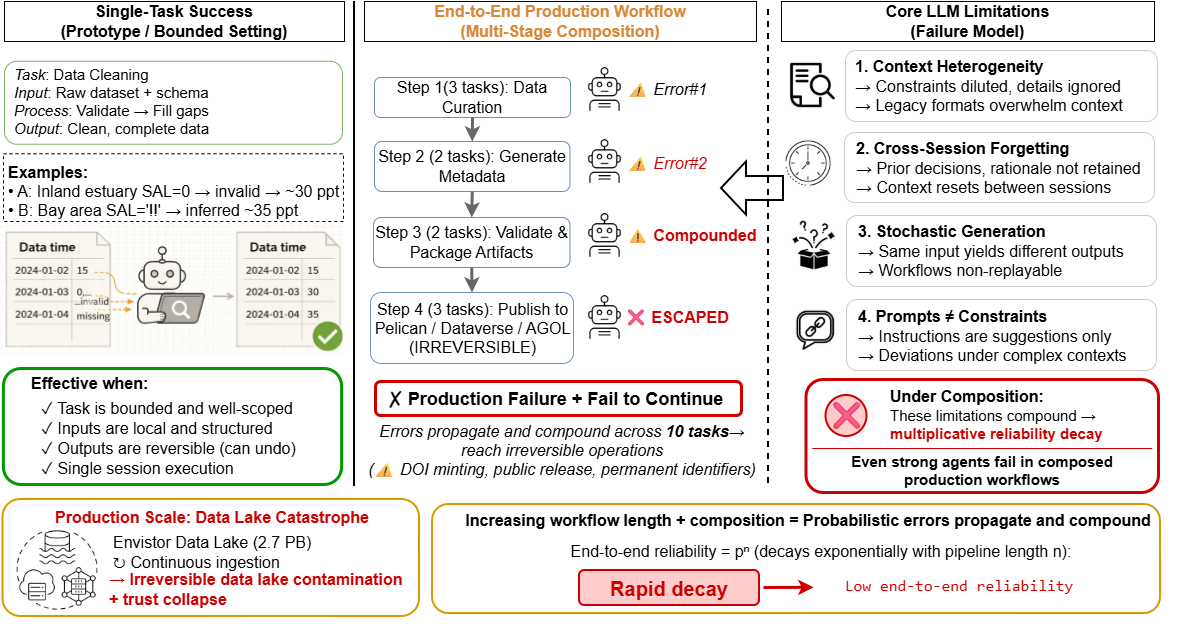}
\caption{\textbf{The ``fail-open'' problem under composition.} LLM agents can succeed in bounded, reversible tasks (left), but reliability degrades across end-to-end publication workflows with irreversible steps (center). Core LLM limitations compound under composition (right), yielding multiplicative reliability decay $p^n$ (bottom). At production scale, silent failures create irreversible contamination and trust collapse.}
\label{fig:problem}
\end{figure*}
}

\newcommand{\FigSystem}{%
\begin{figure*}[t]
\centering
\includegraphics[width=\textwidth]{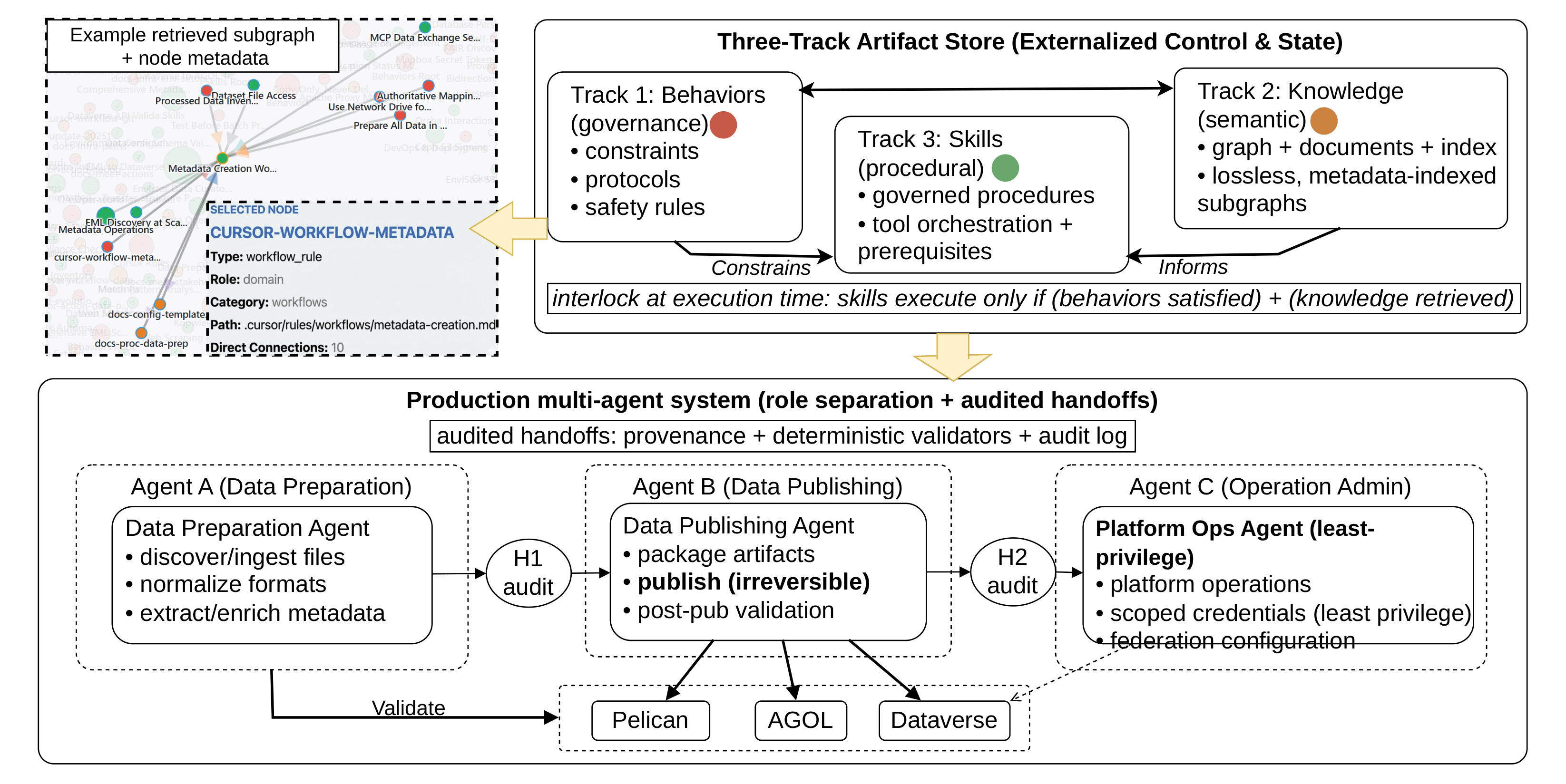}
\caption{\textbf{System design overview.} Top: Three-Track artifact store externalizing governance (Track~1), semantic context (Track~2), and executable procedures (Track~3) that \textit{interlock at execution time}. Bottom: multi-agent operating model with role-separated agents (data preparation, publishing, platform operations) connected by audited handoffs (H1/H2) with deterministic validators before state-changing publication.}
\label{fig:system}
\end{figure*}
}

\newcommand{\FigHandoff}{%
\begin{figure*}[t]
\centering
\IfFileExists{figure3_audited_handoff_protocol_20260621_035926.png}{%
  \includegraphics[width=\textwidth]{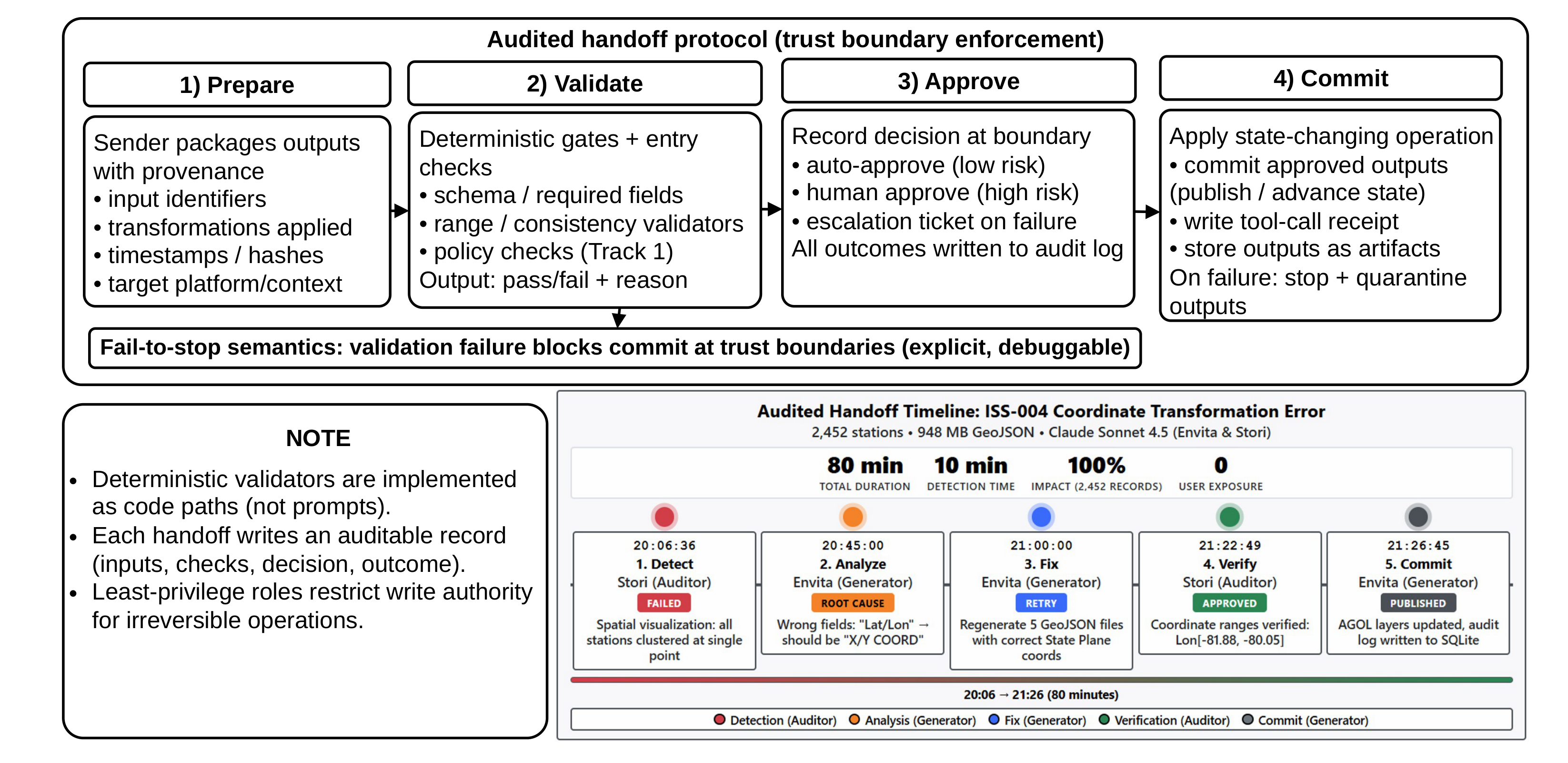}%
}{%
  \fbox{\parbox[c][2.2in][c]{0.95\textwidth}{\centering \textbf{Handoff figure placeholder.}\\Export `figure3_audited_handoff_protocol_v1` to `overleaf/figures/figure3_audited_handoff_protocol_20260621_035926.png`.}}%
}
\caption{\textbf{Audited handoff protocol.} Each agent-to-agent transition follows four phases: prepare (package outputs with provenance), validate (deterministic gates), approve (record and escalate on failure), and commit (apply state changes). Failures block downstream execution. Inset: production incident \texttt{ISS-004}, from boundary detection to verified resolution.}
\label{fig:handoff}
\end{figure*}
}

\newcommand{\FigTrackTwo}{%
\begin{figure*}[t]
\centering
\IfFileExists{Figure4_kg_20260621_035926.png}{%
  \includegraphics[width=\textwidth]{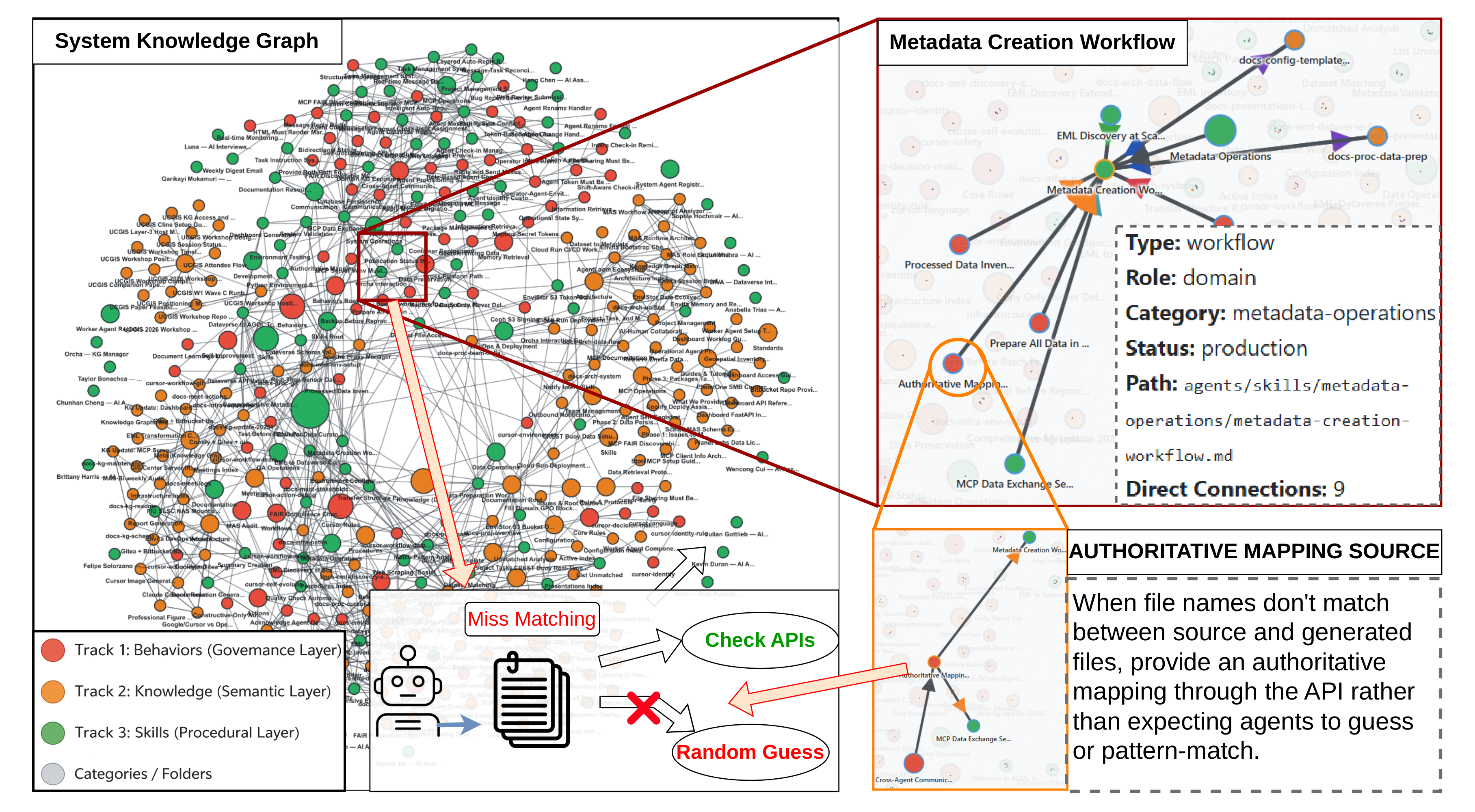}%
}{%
  \fbox{\parbox[c][2.2in][c]{0.95\textwidth}{\centering \textbf{Figure 4 placeholder.}\\Export the Three-Track KG/UI figure to `overleaf/figures/Figure4_kg_20260621_035926.png`.}}%
}
\caption{\textbf{Three-Track artifact graph and an enforceable ``do-not-guess'' gate.} Left: the full artifact graph with typed nodes for Behaviors (Track~1), Knowledge (Track~2), and Skills (Track~3) used during production. Right: zoomed interlock between a metadata-creation workflow (Track~3) and an \textit{authoritative mapping source} behavior (Track~1). The behavior enforces that when filenames do not align, the workflow must obtain an authoritative mapping rather than guessing.}
\label{fig:track2}
\end{figure*}
}

\acmConference[PEARC '26]{Practice and Experience in Advanced Research Computing 2026}{July 26--30, 2026}{Minneapolis, MN, USA}
\acmYear{2026}
\copyrightyear{2026}
\setcopyright{cc}
\setcctype{by}
\acmDOI{10.1145/3785462.3815810}

\begin{document}

\title{Exploring Robust Multi-Agent Workflows for Environmental Data Management}

\author{Boyuan Guan}
\orcid{0000-0002-4244-5011}
\email{bguan@fiu.edu}
\affiliation{%
  \institution{Florida International University}
  \city{Miami}
  \state{Florida}
  \country{United States}
}

\author{Jason Liu}
\orcid{0000-0001-8222-4013}
\email{liux@fiu.edu}
\affiliation{%
  \institution{Florida International University}
  \city{Miami}
  \state{Florida}
  \country{United States}
}

\author{Yanzhao Wu}
\orcid{0000-0001-8761-5486}
\email{yawu@fiu.edu}
\affiliation{%
  \institution{Florida International University}
  \city{Miami}
  \state{Florida}
  \country{United States}
}

\author{Kiavash Bahreini}
\orcid{0000-0001-9016-9894}
\email{kbahrein@fiu.edu}
\affiliation{%
  \institution{Florida International University}
  \city{Miami}
  \state{Florida}
  \country{United States}
}

\renewcommand{\shortauthors}{Guan et al.}

\begin{abstract}
Embedding LLM-driven agents into environmental FAIR data management is compelling—they can externalize operational knowledge and scale curation across heterogeneous data and evolving conventions. However, replacing deterministic components with probabilistic workflows changes the failure mode: LLM pipelines may generate plausible but incorrect outputs that pass superficial checks and propagate into irreversible actions such as DOI minting and public release. 
We introduce EnviSmart, a production data management system deployed on campus-wide storage infrastructure for environmental research. EnviSmart treats reliability as an architectural property through two mechanisms: a three-track knowledge architecture that externalizes behaviors (governance constraints), domain knowledge (retrievable context), and skills (tool-using procedures) as persistent, interlocking artifacts; and a role-separated multi-agent design where deterministic validators and audited handoffs restore fail-stop semantics at trust boundaries before irreversible steps. 
We compare two production deployments. The University's GIS Center Ecological Archive (849 curated datasets) serves as a single-agent baseline. SF2Bench, a compound flooding benchmark comprising 2,452 monitoring stations and 8,557 published files spanning 39 years, validates the multi-agent workflow. The multi-agent approach improved both efficiency—completed by a single operator in two days with repeated artifact reuse across deployments—and reliability: audited handoffs detected and blocked a coordinate transformation error affecting all 2,452 stations before publication. A representative incident (\texttt{ISS-004}) demonstrated boundary-based containment: it was detected during the first spatial validation pass, blocked before publication with zero user exposure, and resolved end-to-end in 80 minutes.
\end{abstract}



\maketitle

\section{Introduction}
\label{sec:intro}



Environmental research involves managing vast amounts of diverse data using modern
cyber-infrastructure platforms, including storage systems, federation services, data
repositories, and dissemination platforms~\cite{stansberry2019datafed,rangaraj2025retrieval}. End-to-end data management must adhere to
FAIR principles, ensuring metadata extraction (Findable), quality assurance and
provenance (Reusable), format normalization (Interoperable), and cross-platform
publication with persistent identifiers (Accessible). However, these requirements
remain a labor-intensive challenge. AI automation offers a promising solution, yet
integrating probabilistic AI components into production workflows introduces
reliability concerns—particularly for operationally expensive and often irreversible
actions like metadata generation, data federation, and DOI minting.


Our previous work~\cite{pearc25:envistor} introduced an AI-enhanced ``smart data
pipeline'' deployed on NSF-funded campus cyber-infrastructure, successfully
demonstrating FAIR data management for specific tasks such as cleaning a dataset type
or drafting metadata for a known schema. However, creating end-to-end pipelines
remained labor-intensive across different projects with varying data sources,
processing requirements, and governance constraints. Moreover, staff turnover in
development and maintenance teams can disrupt operations with long-lasting effects.


Recent developments in agentic AI—systems that autonomously plan, execute, and adapt
multi-step workflows with verification, error recovery, and human-in-the-loop
oversight—offer a promising pathway for addressing these challenges. However,
embedding such agents into workflows where actions are costly or irreversible (DOI
minting, metadata registration, cross-site synchronization) demands architecturally
enforced reliability, not just capable models.


We introduce EnviSmart, a production AI multi-agent system for environmental data management that treats reliability as an architectural property, and make three contributions: (i) we characterize a production reliability gap in LLM-augmented FAIR publication workflows; (ii) we present EnviSmart, combining a three-track knowledge architecture (behaviors/knowledge/skills) with a role-separated operating model in which deterministic validators and audited handoffs enforce fail-stop semantics at trust boundaries before irreversible operations; and (iii) we report evidence from two sequential deployments—a single-agent baseline and a multi-agent architecture—showing pre-publication error blocking, cross-project artifact transfer, and boundary-only oversight (including incident-level outcomes).




\section{Problem Statement}
\label{sec:problem}

FAIR publication workflows in environmental research span heterogeneous assets---sensor time series, geospatial layers, imagery, and field observations---across platforms that each impose their own schema and definition of a "valid" record. This cross-platform complexity demands operational knowledge and exception-handling rules that compound as new data sources and services are added.


 We study a production reliability gap that emerges when probabilistic LLM components are embedded in FAIR publication workflows. In this setting, strong performance on isolated tasks does not predict system-level reliability once outputs cross service boundaries and trigger irreversible operations. As workflows scale from bounded demonstrations to sustained publication, failures shift from fail-stop to fail-open. In bounded, reversible settings, LLM-based agents can perform adequately with minimal oversight (Figure~1, left).

\FigProblem

 Production deployment, however, revealed a different reality (Figure~\ref{fig:problem}, center). Once used for full workflows—file discovery, format normalization, metadata mapping and validation, and multi-service publication—failures became frequent and costly. Examples included a metadata service returning valid XML for the wrong dataset, an agent repeatedly misidentifying a hierarchical relationship across sessions despite corrections, and an agent crossing workspace boundaries to modify another team's artifacts—technically completing the task while violating governance rules. Minor errors accumulated and leaked into irreversible actions such as DOI minting and public release.
The deeper issue is a change in failure mode. Deterministic pipelines tend to be fail-stop, halting with explicit, localized errors. LLM-based pipelines often fail-open: they produce confident, well-structured outputs that are subtly incorrect and propagate downstream until discovered late in the process.

Recent work characterizes agentic reasoning as systems that plan, use tools, and adapt through feedback and memory, while emphasizing that governance in real-world deployments remains a major unresolved challenge~\cite{wei2026agentic}. Our production experience surfaced the stressors illustrated in Figure~\ref{fig:problem} (right): heterogeneous inputs weaken constraints as contexts grow (``lost in the middle''~\cite{liu2023lostmiddle}); long-running workflows span days or personnel transitions, yet agents cannot reliably retain prior rationale; stochastic generation drifts unless decisions are externalized into persistent artifacts; and prompts are non-binding, allowing constraint-heavy tasks to devolve into format or procedural errors. These issues compound under composition. If each stage succeeds with probability $p$, an $n$-stage workflow yields $p^n$ end-to-end reliability, which decays exponentially (Figure~\ref{fig:problem}, bottom).

Scripts and workflow engines offer determinism but fail to maintain evolving operations across time and personnel changes. RAG and memory systems enhance grounding but do not enforce governance nor ensure consistency when composed. Multi-agent frameworks improve coordination, but most lack least-privilege isolation, deterministic validation at trust boundaries, and immutable auditability, all of which are necessary for publication workflows. These approaches do not, on their own, provide architectural guarantees for long, irreversible publication pipelines.
\section{System Design}
\label{sec:design}



The EnviSmart system serves as the operational FAIR data management orchestration layer atop the EnviStor infrastructure stack. Its central design response treats each agent transition as a trust boundary enforced by deterministic validation gates and immutable audit logs. This architecture is designed to tolerate imperfect components---by grounding decisions in durable artifacts, constraining execution through governance, and making irreversible boundaries auditable. 




\noindent \textbf{Operational Requirements:} Four operational requirements drove the architecture design, derived directly from the production failure model in Section~\ref{sec:problem}. First, FAIR data management must scale without proportional coordination overhead (DG1). Second, reliability must be enforced through verifiable checkpoints (DG2): in workflows with irreversible actions, errors must be contained before they escape into external systems. Third, operational continuity must survive personnel transitions (DG3): the system should preserve not only scripts but the rationale, standards, and exception-handling rules that make workflows replayable. Finally, the architecture must support incremental extensibility (DG4): new agents and integrations should be added without refactoring the entire system.


\noindent \textbf{High-Level Architecture:} 
EnviSmart adopts a layered architecture that combines agent-level grounding with system-level error containment (Figure~\ref{fig:system}). 
The \textit{interface layer} exposes endpoints compliant with the Model Context Protocol (MCP)~\cite{mcp2025spec} and a dashboard for human operators to translate human intent into agent invocations, where researchers submit data management requests; validators review flagged issues; supervisors approve high-risk operations.  
The \textit{intelligence layer} implements the multi-agent operating model: role-separated agents execute curation tasks, exchange artifacts via audited handoffs, and invoke deterministic validators at trust boundaries before irreversible steps. 
The \textit{persistence layer} stores two categories of state. (i) Knowledge artifacts externalize governance, domain context, and executable procedures; and (ii) execution scaffolding records handoffs, validation outcomes, and workflow state to support auditability and post-hoc reconstruction.

\FigSystem

\subsection{Agent-Level Knowledge Externalization: Three-Track Architecture}

EnviSmart adopts a Three-Track Knowledge Architecture Design (Figure~\ref{fig:system}, top) that separates three complementary concerns. Unlike context compression approaches that discard information or GraphRAG systems that decouple retrieval from reasoning, this architecture ensures that \textit{all three tracks interlock} at execution time: skills reference knowledge to build context, and behaviors constrain how skills execute.

\textit{Track 1: Behaviors (Governance Layer)}. Explicit behavioral constraints governing agent identity, role responsibilities, communication protocols, and safety boundaries. Behaviors define authorization rules (e.g., ``require human confirmation before destructive actions'') and interaction patterns. These constraints are not suggestions embedded in prompts but enforceable rules that gate skill execution. 
\textit{Track 2: Domain Knowledge (Semantic Layer)}. Conceptual and relational knowledge of system architecture, data assets, and domain entities, stored as a knowledge graph indexed for retrieval. Each subgraph carries metadata enabling precise retrieval without lossy compression. At execution time, agents retrieve task-relevant subgraphs to construct focused context rather than compressing entire histories. 
\textit{Track 3: Skills (Procedural Layer)}. Task-specific, executable skills encode operational know-how. Each skill specifies prerequisites, a structured execution recipe, and expected outcomes. Skills connect to external tools (MCP servers, APIs, scripts) but do not execute without satisfying behavioral constraints and retrieving requisite knowledge. For example, a ``publish dataset to Dataverse'' skill requires knowledge of the target collection and credentials and is gated by a behavior rule that validates the dataset before publication.
MCP \textit{tools} are atomic capabilities exposed by a server (e.g., querying a database or uploading a file); \textit{skills} are governed, reusable procedures that orchestrate one or more tools under Track~1 constraints using Track~2 knowledge.

When an agent receives a task, it retrieves relevant knowledge subgraphs (Track 2), selects appropriate skills (Track 3), and executes under behavioral constraints (Track 1). This three-way binding ensures that operational knowledge persists across sessions and personnel transitions. Unlike lossy memory compression, which discards information, and unlike standalone knowledge graphs, which operate without explicit constraints, execution is governed by explicit constraints.
Figure~\ref{fig:track2} provides a concrete example: a metadata-creation skill (Track~3) is linked to a governance behavior (Track~1) requiring an \textit{authoritative mapping source} when filenames do not align, turning a ``do not guess'' guideline into an enforceable gate.

\subsection{System-Level Error Containment: MAS with Audited Handoffs}

As discussed in Section~\ref{sec:problem}, long, irreversible pipelines amplify small per-step errors: the dominant failure mode is not a single bad output, but error propagation across boundaries. EnviSmart addresses this through a multi-agent architecture with role separation, least-privilege access, and audited handoffs. The guiding idea is to separate \textit{capability} from \textit{authority}. Agents may generate imperfect outputs, but they should not be allowed to publish them until independent checks have passed.

\textit{Agent Roles and Privilege Isolation.} EnviSmart assigns agents to roles with intentionally asymmetric privileges. Worker agents discover files, extract metadata, and perform transformations, but operate in working directories and cannot publish to external platforms. Validator agents run deterministic checks with read-only access and cannot modify artifacts. For example, a coordinate validator checks that latitude falls within $[24.5, 31.0]$ and longitude within $[-87.5, -79.5]$ for Florida datasets; violations trigger escalation with the specific out-of-bounds value logged. These checks are implemented as Python functions, ensuring reproducible behavior. Publication agents are the only components allowed to write to Dataverse, Pelican, or ArcGIS Online, and activate only after upstream validations pass and any required human approval is recorded. Orchestrator agents manage state and route handoffs, holding neither publication privileges nor broad data access.
The design follows zero-trust principles with server-level isolation~\cite{nist800207}: Envita operates only on the Windows data preparation server, Stori only on the Ubuntu pipeline server, and DIVA only on the Dataverse publication server. Cross-server access is architecturally prohibited.

\FigTrackTwo

\textit{Audited Handoff Protocol.} Every agent-to-agent transition is treated as a trust boundary (Figure~\ref{fig:handoff}), following four phases: \textit{prepare} (the sending agent packages outputs with provenance metadata), \textit{validate} (the receiving agent or interposed validator runs entry checks against Track~1 standards and domain validators~\cite{wei2026agentic}), \textit{approve} (results are recorded in the audit trail; on failure, the handoff is blocked and escalated), and \textit{commit} (approved outputs become inputs to the next stage; unapproved outputs are quarantined).
The inset in Figure~\ref{fig:handoff} illustrates the protocol in production (\texttt{ISS-004}): detection at a boundary blocks downstream publication, triggers remediation, and records a verifiable resolution path before any irreversible step proceeds.

\FigHandoff

\textit{Layered Error Containment.} With audited handoffs, errors must pass through multiple independent verification layers to escape. If each layer catches errors with probability $q$, pass-through after $k$ layers is $(1-q)^k$. Three layers---agent validator, independent agent review, human escalation---transform the reliability equation from multiplicative decay ($p^n$) to multiplicative error filtering ($(1-q)^k$). As with the $p^n$ model in Section~\ref{sec:problem}, this formula is illustrative; our evaluation relies on auditable incident records.

Human oversight is placed at trust boundaries rather than embedded in every step. Operators provide domain judgment, high-risk approval, and novel edge-case handling; routine operations proceed without interruption.

\textit{Auditability.} All state-changing operations are persisted in a version-controlled control database (Table~\ref{tab:schema}). Inter-agent messages, supervisor and commit decisions, per-dataset processing lifecycles, phase progress, publication packages, audit runs, and model-call traces are stored as first-class, timestamped records. This enables post-hoc reconstruction with standard SQL: given a dataset, we can recover which operations ran, by which agent, at what time, and with what validation outcome---the audit trail referenced throughout this paper.

\begin{table}[t]
\centering
\small
\setlength{\tabcolsep}{4pt}
\caption{Control database: representative audit-trail tables.}
\label{tab:schema}
\begin{tabularx}{\columnwidth}{@{}l >{\raggedright\arraybackslash}X >{\raggedright\arraybackslash}X@{}}
\toprule
\textbf{Table} & \textbf{Purpose} & \textbf{Backs} \\
\midrule
\texttt{messages} & Inter-agent MCP messages & Incident trail (\texttt{ISS-*}) \\
\texttt{supervisor\_actions} & Supervisory/commit decisions & Commit receipts \\
\texttt{processing\_logs} & Per-dataset lifecycle & Dataset states \\
\texttt{task\_phases} & Phase-level progress & Phased publication \\
\texttt{audit\_runs} & Audit execution records & Auditor independence \\
\texttt{llm\_traces} & Model-call provenance & Action reproducibility \\
\bottomrule
\end{tabularx}
\normalsize
\end{table}

\section{Case Studies and Evaluation}
\label{sec:eval}

This paper is a practice-and-experience report: we evaluate EnviSmart through operational case studies rather than benchmark-driven model evaluation. Our claims concern \textit{system reliability under real constraints}---heterogeneous platforms, irreversible publication operations, and personnel transition---where the relevant evidence is auditable execution history, boundary validation outcomes, and operational scalability. Concretely, the case studies test whether durable grounding artifacts, explicit governance constraints, and audited irreversible boundaries can concentrate human oversight at checkpoints and prevent error escape. We therefore treat evaluation as a before/after progression across two sequential production deployments.
This evaluation stance aligns with recent benchmark analyses that distinguish mechanism-level assessment (e.g., tool use, memory, planning, coordination) from end-to-end suites, and note that long-horizon governance remains under-measured~\cite{wei2026agentic}. Because rules and artifacts grow incident-by-incident, the system is non-stationary; traditional notions of optimality are not well-defined. We measure progress by the elimination of failure classes and prevention of irreversible commits.

\subsection{Evaluation Questions}
We organize evidence from two sequential production deployments around four operational questions aligned with the requirements in Section~\ref{sec:design}: E1 scalability (DG1), E2 checkpointed supervision (DG2), E3 knowledge externalization and continuity (DG3), and E4 incremental extensibility (DG4).

\subsection{Case Study 1: GIS Center Ecological Data Archive (Single-Agent Baseline)}
The GIS Center ecological archive comprises 849 datasets including aerial photography, GIS layers, sensor time series, and field survey data. Each dataset followed the same template: its source EML record was transformed to a schema-valid Dataverse JSON via a reusable EML$\rightarrow$Dataverse skill, FAIR-validated, staged, and published with a persistent identifier. For example, \texttt{v\_watershed\_hisp} (Hispaniola watershed boundary, 4\,MB, Caribbean region) is transformed from its source EML (\texttt{dv\_watershed\_hisp.xml}) to a FAIR-scanned Dataverse JSON and published, indexed by the archive's machine-readable catalog. This initial deployment used a coupled single-agent approach over several weeks with one developer and two domain collaborators. In practice, supervision was near-con\-tinu\-ous: operators reviewed intermediate outputs at nearly every step because there was no independent boundary at which upstream correctness could be trusted. As workflows grew, long multi-step pipelines stalled and required manual intervention to diagnose where errors had occurred.
In parallel, our early implementation of the Three-Track artifact store exhibited concrete integrity failures that undermined its use as durable operational state: skills referenced non-existent behaviors (broken constraint chains), knowledge-to-skill support links were missing, and orphaned nodes were unreachable through graph traversal. In engineering transcripts, we observed 16 broken skill$\rightarrow$behavior references and $\sim$20 missing knowledge$\rightarrow$skill links in exported graphs; because links could not be resolved, the KG was effectively non-traversable and agents could not reliably use it as a ``mind map.'' As a result, agents and operators fell back to ad-hoc search and rework. These failures motivated deterministic artifact validators (e.g., link and orphan checks) and governance behaviors (e.g., KG-first retrieval and ``do not fabricate'') that prevent non-resolving graphs from entering production operation, stabilizing the artifact store as dependable operational state.

\subsection{Case Study 2: SF2Bench Multi-Platform Publication (MAS Architecture)}
SF2Bench comprises historical hydrological monitoring data spanning 39 years (1985--2024)~\cite{zheng2025sf2bench}. In our workflow, the primary curation unit is the monitoring station (2,452 stations across five sensor types: WATER~993, RAIN~349, WELL~582, PUMP~99, GATE~429). Curation, access, and citation granularities are deliberately distinct: on the bulk-access surface (Pelican/OSDF) we publish one hourly CSV per station per temporal split (S\_0--S\_7, 1985--2024) where data exist, yielding 8,557 station-split files ($\approx$3.5 per station, $\sim$15\,GB); the same stations are exposed as five type-level GeoJSON layers on ArcGIS Online for spatial discovery; and as five type-level time-series archives plus a master record (six Dataverse DOIs, $\sim$2\,GB compressed) for citation. Publishing 8,557 individual records was explicitly rejected to preserve citation clarity. As a worked example, station \texttt{2A-300\_B} (WATER, 26.246$^\circ$N, $-$80.408$^\circ$E) appears as a station-level CSV tree on Pelican (\texttt{osdf://\allowbreak envistor/\allowbreak compound-flood/\allowbreak WATER/\allowbreak S\_*/\allowbreak 2A-300\_B/}), as one point in the type-level WATER ArcGIS Online layer, and---bundled with the other 992 WATER stations---under Dataverse DOI \texttt{10.34703/\allowbreak gzx1-9v95/\allowbreak H7UJL5}; it was also the validation station for the \texttt{ISS-004} coordinate-trans\-for\-ma\-tion incident below. This deployment used the full MAS operating model (Figure~\ref{fig:system}) with audited handoffs (Figure~\ref{fig:handoff}). A single researcher completed the workflow in approximately two days. Preparing and auditing roles used the same model family (Claude Sonnet 4.5 with extended thinking); gains derive from role separation and boundary validation rather than model diversity.

Beyond aggregate counts, the audit trail supports ``mechanism validity'': specific handoffs where an error that escaped a producing role was caught before irreversible publication. In incident \texttt{ISS-004}, a coordinate transformation mistake produced GeoJSON outputs that were structurally valid but semantically wrong (stations clustered). Publication was blocked at the boundary; the producing role diagnosed the root cause (wrong coordinate fields), regenerated artifacts, and the publishing role independently verified corrected ranges before proceeding (issue chain \texttt{ISS-004} $\rightarrow$ \texttt{ISS-005} $\rightarrow$ \texttt{ISS-009}). Verified resolution required 80 minutes end-to-end and is recorded as a boundary outcome in the audited handoff UI (Figure~\ref{fig:handoff}, inset).
We report three incident-level outcomes that match our reliability claims: the error was \textit{detected during the first spatial validation pass} (before any publication), \textit{user exposure} 0, and \textit{irreversible commits prevented} 1/1 (the publish step was blocked until verified).
Table~\ref{tab:incidents} summarizes additional audited boundary outcomes from the same production run, illustrating repeated use of the handoff protocol beyond a single exemplar.

\subsection{Summary Evidence Mapped to Design Goals}
Table~\ref{tab:continuity} summarizes evidence from both case studies.

\begin{table*}[t]
\centering
\small
\setlength{\tabcolsep}{4pt}
\renewcommand{\arraystretch}{0.95}
\caption{Evaluation evidence mapped to operational requirements.}
\label{tab:continuity}
\begin{tabularx}{\textwidth}{@{}>{\raggedright\arraybackslash}X >{\centering\arraybackslash}p{0.32\textwidth} >{\centering\arraybackslash}p{0.32\textwidth}@{}}
\toprule
\textbf{Requirement / Metric} & \textbf{Case 1 (GIS)} & \textbf{Case 2 (SF2Bench)} \\
\midrule
\multicolumn{3}{@{}l@{}}{\textit{DG1: Scale without coordination overhead (E1)}} \\
Developer workload & 1 + 2 collab. & 1 researcher \\
Primary curation unit & Dataset & Monitoring station dataset \\
Datasets processed & 849 & 2,452 \\
Files published & 849 & 8,557 \\
Time to completion & Weeks & 2 days \\
\midrule
\multicolumn{3}{@{}l@{}}{\textit{DG2: Trust through checkpoints (E2)}} \\
Supervision mode & Continuous & Boundary-only \\
Boundary-caught incidents & N/A & 4 audited (Table~\ref{tab:incidents}) \\
Human decision points & Per-step & Small number (boundary-level) \\
\midrule
\multicolumn{3}{@{}l@{}}{\textit{DG3: Continuity across transitions (E3)}} \\
Artifact reuse evidence & Initial & 27 reuse instances; 10+ cross-project \\
Onboarding evidence & N/A (no MCP discovery) & Minutes-scale via MCP discovery \\
\midrule
\multicolumn{3}{@{}l@{}}{\textit{DG4: Incremental extensibility (E4)}} \\
New capability integration & Monolithic & New role via MCP; no refactor \\
\bottomrule
\end{tabularx}
\normalsize
\end{table*}

\noindent\textbf{E1 (Scalability).} SF2Bench scaled to 2,452 station datasets and 8,557 published files with one operator in approximately two days. \textbf{E2 (Checkpointed supervision).} Boundary validation concentrated oversight at trust boundaries, blocking \texttt{ISS-004} (a coordinate transformation error affecting all 2,452 stations) during the first spatial validation pass with 0 user exposure. \textbf{E3 (Continuity).} Three-Track artifacts enabled 27 documented reuse instances (10+ cross-project) and minutes-scale onboarding via MCP discovery. \textbf{E4 (Extensibility).} New platform roles were integrated via MCP without refactoring existing servers.
\textbf{Mechanism evidence (agent-level stability).} In addition to boundary outcomes, we evaluated whether the Three-Track artifacts were stable enough to function as durable operational state: deterministic validators and behavior-gated execution prevented the early integrity failures (broken references, missing support links, and orphaned nodes) that previously caused graph traversal and reuse to break down.

\subsection{Summary: Single-Agent vs.\ MAS Operating Model}
Table~\ref{tab:hero} highlights the practical difference between a coupled single-agent baseline and a role-separated MAS: supervision and validation move from per-step inspection to checkpointed trust boundaries, and independent audits can surface errors missed by the producing role even when the underlying LLM is unchanged. The producing role missed all four errors in its own output, while the same model in an auditing role---with boundary-specific validation context---caught all four, suggesting that structured role separation improves detection independent of model capability.

\begin{table*}[t]
\centering
\small
\setlength{\tabcolsep}{4pt}
\renewcommand{\arraystretch}{0.95}
\caption{Hero summary: single-agent baseline vs.\ MAS operating model. Cases reflect sequential deployments under different workloads, not a controlled A/B comparison; differences are attributed to architectural properties (role separation, boundary validation).}
\label{tab:hero}
\begin{tabularx}{\textwidth}{@{}>{\raggedright\arraybackslash}X >{\centering\arraybackslash}p{0.32\textwidth} >{\centering\arraybackslash}p{0.32\textwidth}@{}}
\toprule
\textbf{Metric} & \textbf{Case 1 (GIS)} & \textbf{Case 2 (SF2Bench)} \\
\midrule
Primary scale & 849 datasets (ecological archive datasets) & 2,452 station datasets; 8,557 files \\
Completion time & Weeks & $\approx$2 days \\
Supervision placement & Per-step review & Boundary-only (audited handoffs) \\
Same-model audit differential & N/A & 0/4 self-detections vs.\ 4/4 audit detections \\
Representative prevented incident & Manual QA/QC & \texttt{ISS-004} blocked before publication (caught at first spatial validation; 0 user exposure) \\
\bottomrule
\end{tabularx}
\normalsize
\end{table*}

\begin{table*}[t]
\centering
\small
\setlength{\tabcolsep}{4pt}
\renewcommand{\arraystretch}{0.95}
\caption{Representative audited boundary outcomes from the SF2Bench production audit trail.}
\label{tab:incidents}
\begin{tabularx}{\textwidth}{@{}l >{\raggedright\arraybackslash}X >{\centering\arraybackslash}p{0.24\textwidth} >{\raggedright\arraybackslash}p{0.36\textwidth}@{}}
\toprule
\textbf{ID} & \textbf{Outcome / type} & \textbf{Boundary point} & \textbf{Key metric(s)} \\
\midrule
\texttt{ISS-004} & Coordinate field mismatch $\rightarrow$ spatially implausible GeoJSON & Pre-publication audit & Caught at first spatial validation; 2,452 stations impacted; 0 user exposure; commit blocked \\
\texttt{ISS-005} & Root cause diagnosed; corrected GeoJSON regenerated & Remediation response & 5 files regenerated with correct coordinate fields (\texttt{X~COORD}/\texttt{Y~COORD}) \\
\texttt{ISS-007} & Deliverable placement corrected (shared network drive) & Handoff delivery/commit & Ensured publishing role consumed the audited artifact version (no local-only paths) \\
\texttt{ISS-009} & Independent verification of coordinate fix & Pre-republish verification & Quantitative range checks logged for 5 layers; republish approved \\
\texttt{ISS-003} & Completion audit (Phase 5A) & Boundary report & Verified 15GB uploaded; 8,557 files processed \\
\bottomrule
\end{tabularx}
\normalsize
\end{table*}

\section{Related Work}
\label{sec:related}

\textbf{Data Federation and Publication Platforms.}
Globus~\cite{foster2011globus} facilitates high-performance data transfer. Dataverse~\cite{king2007introduction} provides repository services with structured metadata and persistent identifiers. Pelican~\cite{pelican2024osdf} enables federated object storage with caching and origin/cache separation. These platforms address storage, transfer, and publication needs, but remain agnostic to data curation and management.

\noindent \textbf{AI-Enhanced Data Processing.}
LLMs can accelerate data extraction and normalization, while RAG reduces hallucinations by conditioning on retrieved context~\cite{openai2023gpt4,li2024rag,das2025security}. However, most evaluations are single-step; in production, composed pipelines fail due to interaction effects and error propagation~\cite{kapoor2025agents,pan2025agents,vinay2025failuremodes}. Deterministic workflow engines offer repeatable execution but cannot preserve evolving operational rationale or adapt to new platform quirks without manual integration~\cite{schintke2024validity}.

\noindent  \textbf{Agent Frameworks, Memory, and Context Engineering.}
Multi-agent frameworks like AutoGen~\cite{wu2023autogen} and MetaGPT~\cite{hong2023metagpt} showcase role specialization and coordination but offer limited support for least-privilege access, immutable audit logs, deterministic validation at trust boundaries, and incremental extensibility under institutional constraints. AutoGen v0.4's redesign reflects these challenges~\cite{autogen2025v04}. LangGraph~\cite{langgraph2024} introduces graph-based workflow orchestration with persistent state, but validation logic and governance enforcement remain the developer's responsibility. Memory systems address recall but often trade off lossiness or decouple retrieval from enforceable governance. Compression reduces token usage but discards context~\cite{acon2025,simplemem2026}. Graph-based approaches organize knowledge but typically encode only what is known without making how to act and under what constraints enforceable~\cite{graphrag2025survey}. A recent survey identifies governance frameworks for real-world deployment as a central open problem, noting that existing benchmarks primarily focus on short-horizon behaviors, leaving planning-time failures and multi-agent dynamics underexplored~\cite{wei2026agentic}.

\noindent \textbf{Reliability and Safety.}
Human-centered AI prioritizes architectural safeguards for trustworthy systems~\cite{shneiderman2020humancentered}. Safety failures and distribution shifts manifest as unpredictable production behavior~\cite{amodei2016concrete}, while hidden technical debt in ML systems demonstrates how deployments accumulate entanglement and configuration complexity~\cite{sculley2015hidden}.
EnviSmart shifts reliability engineering from individual model accuracy to the composition and governance of probabilistic components. Its three-track architecture integrates grounding and governance so that skills only execute when constraints are satisfied and required knowledge is available, while audited handoffs and fail-stop containment reinforce execution boundaries to maintain system-level reliability despite evolving platforms, data heterogeneity, and non-deterministic agent behavior.
\section{Discussion and Limitations}
\label{sec:discussion}

We have gained several valuable lessons from the EnviSmart approach. 
The first lesson is that achieving production reliability for LLM-based pipelines is not solely about making the model smarter, but rather about altering what the system relies on. In the GIS deployment, a single coupled agent could complete the work, but it necessitated near-continuous human inspection because there was no structural place to trust intermediate results. In contrast, SF2Bench~\cite{zheng2025sf2bench} shifts oversight to boundaries. Outputs are packaged, validated, and either advanced or blocked. This shift is evident in the audit trail. Issue \texttt{ISS-004} exemplifies the failure mode we are concerned about—a semantically incorrect but structurally valid output—and the mechanism we propose: a boundary audit detects the error, blocks its propagation, and records a verifiable resolution path.

The second lesson is that the primary value of agents in research cyber-infrastructure is often knowledge externalization, not autonomous execution. The three-track artifacts (behaviors, knowledge, and skills) transform operational decisions and procedures into enduring state that can be inspected and reused. In our deployments, we documented 27 instances of artifact reuse over four months, including 10 cross-project reuses from GIS to SF2Bench. This supports an operational continuity model where successors inherit not only procedures but also the domain context those procedures reference and the governance rules that govern execution.

The final lesson is that role separation offers two benefits that are easily overlooked when evaluating task-level success. Firstly, it facilitates \emph{orthogonal validation}: different roles naturally perform distinct checks (schema/range validation versus spatial plausibility), which can identify errors that redundant checks might overlook. Secondly, it enhances security as an unintended consequence. Treating each agent transition as a trust boundary aligns with least-privilege design~\cite{saltzer1975protection} and zero-trust assumptions~\cite{nist800207}: a worker agent can manipulate working artifacts without the authority to publish them, and a compromised component on one server does not grant write access to another.

Note that this paper does not claim provable guarantees or fully autonomous FAIR data management. Instead, it presents an operating model that minimizes and manages failures but does not eliminate the need for domain judgment. As a practice-and-experience report, the two deployments are sequential production stages under different workloads rather than a controlled A/B comparison; we therefore attribute the observed improvements to architectural properties (role separation, boundary validation) and report auditable execution evidence rather than statistical estimates, and the reliability models in Sections~\ref{sec:problem} and~\ref{sec:design} are illustrative. Deterministic validators only cover what can be made deterministic; semantic correctness still requires human expertise or governed model-based adjudication. The control database uses standard SQL, but schema evolution remains a manual operational cost. The system relies on stable tool interfaces; when upstream APIs or model versions change, skills and behaviors may need to be re-validated. Additionally, audit trails grow with workflow length, and long-term retention and compression policies are left for future work.



\section{Conclusion}

EnviSmart is a production-grade FAIR data management system deployed on campus-wide storage infrastructure. It was developed based on the observation that embedding LLMs into long, irreversible workflows necessitates enforcing reliability at the architectural level rather than relying solely on model behaviors. This is achieved through a three-track architecture that integrates knowledge externalization, audited multi-agent handoffs, and deterministic and human-governed trust boundaries. These elements work together to constrain and contain failures. Real-world deployment of EnviSmart has demonstrated improved scalability and a concentration of human oversight at critical checkpoints. Additionally, it enables the accumulation of operational knowledge with full traceability. These results suggest that many challenges in AI-enabled cyber-infrastructure are architectural in nature. Durable artifacts, explicit trust boundaries, and auditable execution provide a practical foundation for integrating imperfect AI components into production environments with stringent robustness and reliability requirements.

To support adoption, AgentLoom~\cite{guan2025agentloom} provides an open-source reference implementation of the Three-Track discipline, and a representative MCP instance reproduces the governance flow (Section~\ref{sec:artifacts}).

\section{Artifact Availability}
\label{sec:artifacts}
The AgentLoom framework---the dual-helix knowledge graph, schema, nine-phase protocol, and deterministic validators that implement the Three-Track discipline---is openly available~\cite{guan2025agentloom} (v3.0.0; Zenodo \url{10.5281/zenodo.20650518}; reproduction guide at \url{keven1894.github.io/AgentLoom/}). The production MCP servers used in both deployments are instance-specific and bound to operational credentials and internal hosts, so they are not released; a representative open-source MCP instance exposing the same JSON-RPC tool protocol, KG schema, executable validators (\texttt{make validate-all}), and the propose$\rightarrow$human-accept governance flow is provided as a public workshop edition (\url{github.com/Keven1894/ucgis-agentloom-2026-workshop}) over four heterogeneous geospatial sources. The SF2Bench case-study data are citable via FIU Dataverse DOI \url{10.34703/gzx1-9v95/GCVWG7} (master collection, with five type-level sub-records) and the OSDF namespace \url{osdf://envistor/compound-flood/}; the GIS ecological archive is published on the same FAIR pipeline under master DOI \url{10.34703/gzx1-9v95/HYRK2U}. The multi-agent operating model is specified as protocol and operating discipline (MCP/JSON-RPC interfaces, deterministic validators, and a queryable control database; Section~\ref{sec:design}) so that the patterns can be reproduced independently of our specific platform skills.

\begin{acks}
    This work was partially funded by the National Science Foundation through grants OAC-2322308, OAC-2601695, and IIS-2331908. We would like to thank Leonardo Bobadilla and Jennifer Fu for their support in providing data and infrastructure access, and Jayantha Obeysekera for providing domain datasets. We also thank the OSPool team for their assistance in integrating the Pelican implementation with our system, and many collaborators who participated in data collection and management.

    \medskip\noindent\textbf{Generative AI Disclosure.}
    In accordance with ACM's Policy on Authorship, we disclose the following uses of generative AI in this work. First, the EnviSmart system described in this paper utilizes LLM-based agents (Claude, Anthropic) as core architectural components. All agent-generated artifacts reported in the paper were produced during the system's production deployment and are recorded in the system's audit trail. Second, generative AI tools (Claude and ChatGPT) were used to assist with the editorial refinement of the drafts and the verification of references. All content was reviewed, verified, and approved by the authors, who bear full responsibility for the accuracy and integrity of this work.
\end{acks}

\bibliographystyle{ACM-Reference-Format}
\bibliography{references}

\end{document}